\documentclass{article}

\usepackage{microtype}
\usepackage{graphicx}
\usepackage{subfigure}
\usepackage{booktabs} 

\usepackage{hyperref}

\usepackage[ruled]{algorithm2e}
\usepackage{paralist}
\usepackage{bm}
\usepackage{amsfonts, amsmath}       

\usepackage[accepted]{icml2021}

\icmltitlerunning{Twin Weisfeiler-Lehman: High Expressive GNNs for Graph Classification}

\begin{document}

\twocolumn[
\icmltitle{Twin Weisfeiler-Lehman: High Expressive GNNs for Graph Classification}

\begin{icmlauthorlist}
\icmlauthor{Zhaohui Wang}{label1,label3}
\icmlauthor{Qi Cao}{label1}
\icmlauthor{Huawei Shen}{label1,label3}
\icmlauthor{Bingbing Xu}{label1}
\icmlauthor{Xueqi Cheng}{label3,label2}
\end{icmlauthorlist}
\icmlaffiliation{label1}{Data Intelligence System Research Center, Institute of Computing Technology, Chinese Academy of Sciences}
\icmlaffiliation{label2}{CAS Key Laboratory of Network Data Science and Technology, Institute of Computing Technology, Chinese Academy of Sciences}
\icmlaffiliation{label3}{University of Chinese Academy of Sciences}
\icmlcorrespondingauthor{Huawei Shen}{shenhuawei@ict.ac.cn}

\icmlkeywords{Machine Learning, ICML}

\vskip 0.3in
]



\printAffiliationsAndNotice{}  

\begin{abstract}
The expressive power of message passing GNNs is upper-bounded by Weisfeiler-Lehman (WL) test. To achieve high expressive GNNs beyond WL test, we propose a novel graph isomorphism test method, namely Twin-WL, which simultaneously passes node labels and node identities rather than only passes node label as WL. The identity-passing mechanism encodes complete structure information of rooted subgraph, and thus Twin-WL can offer extra power beyond WL at distinguishing graph structures. Based on Twin-WL, we implement two Twin-GNNs for graph classification via defining readout function over rooted subgraph: one simply readouts the size of rooted subgraph and the other readouts rich structure information of subgraph following a GNN-style. We prove that the two Twin-GNNs both have higher expressive power than traditional message passing GNNs. Experiments also demonstrate the Twin-GNNs significantly outperform state-of-the-art methods at the task of graph classification.
\end{abstract}

\section{Introduction}
Graph neural networks (GNNs) have achieved state-of-the-art performance in graph classification task \cite{wu2020comprehensive,abadal2021computing,zhou2022graph}. The success of GNNs lies in their powerful capacity at graph representation learning, following a message passing paradigm that iteratively aggregates neighbor information and pools node representations into graph-level representations \cite{gilmer2017neural}. However, the expressive power of GNNs is theoretically upper-bounded by Weisfeiler-Lehman (WL) test \cite{xu2018powerful,morris2019weisfeiler}, i.e., a typical graph isomorphism test method. An important and challenging problem is: \emph{ how to design high expressive GNNs that can exceed the WL test?}

\begin{figure}[t]
\vskip 0.2in
\begin{center}
\centerline{\includegraphics[width=1\columnwidth]{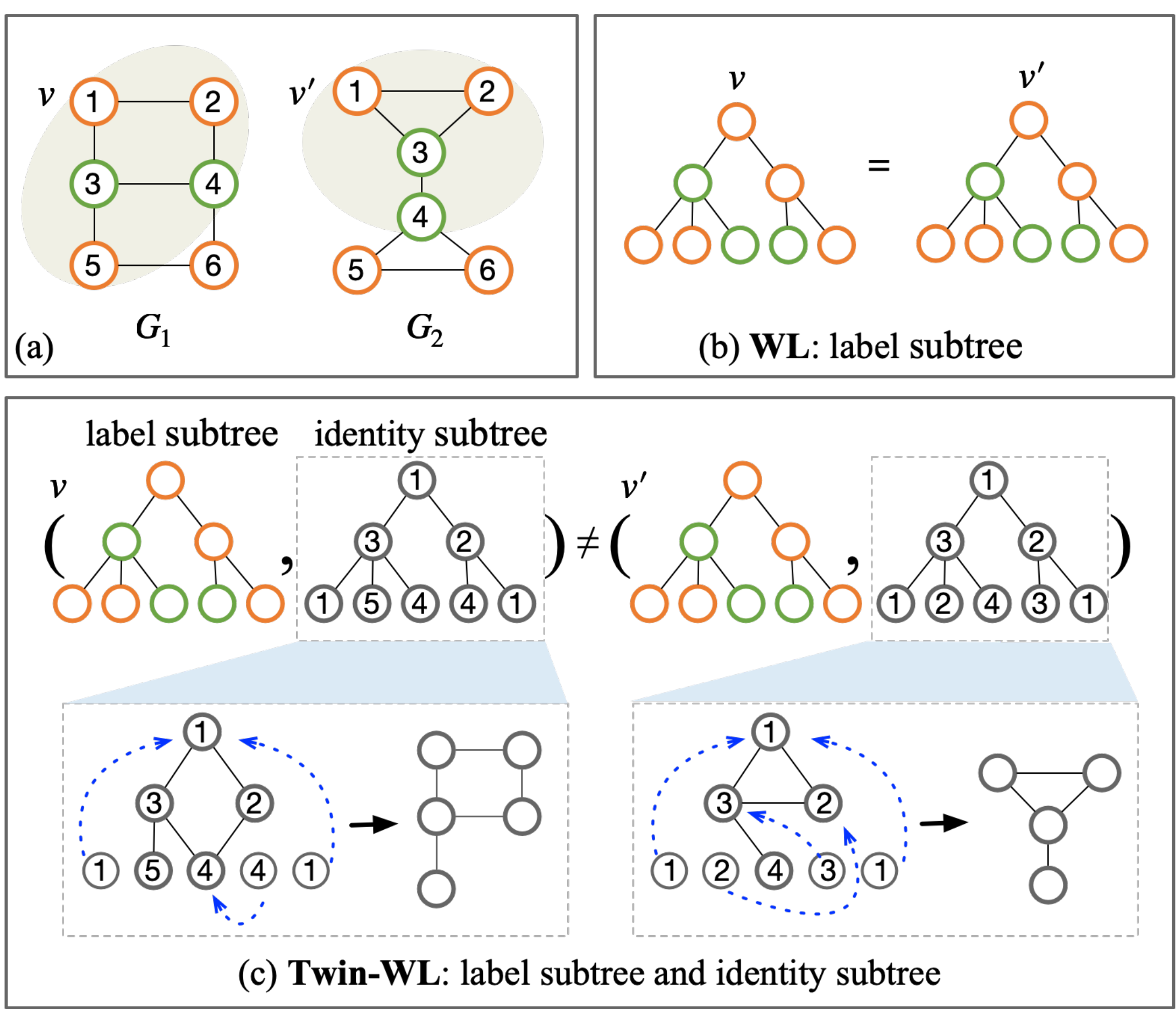}}
\caption{Comparison of Twin-WL with the WL. Given two non-isomorphic graphs (a). WL encodes the node by label subtree. Different rooted subgraphs of $v$ and $v'$ map to the same subtree (b). Twin-WL encodes the node by twin subtrees tuple including a label subtree and an identity subtree. The distinguishing power of Twin-WL is higher than that of WL (c).}
\label{fig:motivation}
\end{center}
\vskip -0.1in
\end{figure}

To achieve high expressive GNNs beyond WL test, we propose a novel graph isomorphism test method, namely Twin Weisfeiler-Lehman (\textbf{Twin-WL}). Twin-WL distinguishes itself from WL by simultaneously passing node labels and node identities, rather than solely passing node labels. The identity-passing mechanism encodes complete structure information of rooted subgraph into an identity subtree. As a result, our proposed Twin-WL can offer extra power beyond WL at distinguishing graph structures as illustrated in Figure~\ref{fig:motivation}. Twin-WL opens a new door to design high expressive GNNs for graph classification. 

Guided by Twin-WL, the key of designing high expressive GNNs is to define a readout function to extract discriminative structure information of the rooted subgraph from the identity subtree. In this paper, we propose two Twin-GNNs via implementing different readout functions in Twin-WL. When the labeled data is inadequate, we define the readout function as the \underline{s}ize of rooted subgraph for graph classification, namely STwin-GNN. When the labeled data is adequate, we utilize a powerful \underline{n}eural network to define the readout function, namely NTwin-GNN.  
   
We evaluate the effectiveness of the proposed two GNNs on graph classification task via several benchmark datasets. The experimental results demonstrate that our methods significantly outperform competitive baseline methods. Moreover, we also conduct the expressive power evaluation, case study, and runtime comparison to comprehensively analyze the effectiveness and the efficiency. 

The main \textbf{contributions} of this paper include: 
\begin{itemize}
\item We propose a novel Twin-WL graph isomorphism test method that achieves beyond-WL ability at distinguishing graph structure, opening a new door for designing higher expressive GNNs.
\item Based on Twin-WL, we design two Twin-GNNs, possessing high expressiveness beyond traditional message passing GNNs and achieving significant improvements on graph classification.
\end{itemize}

\section{Preliminary}
\subsection{Graph Isomorphism}
We define a graph as $G=(V,E)$, where $V$ and $E$ are the sets of nodes and edges respectively. Two graphs $G$ and $G'$ are isomorphic if there exists a bijection $\xi$ between $V$ and $V'$. $\xi: V\rightarrow V'$ and it preserves the edge relation, i.e., $(u,v)\in E$ if and only if $\left(\xi(u),\xi(v)\right)\in E'$ for all $u,v\in V$. 

\subsection{Weisfeiler-Lehman Test}
 Weisfeiler-Lehman (1-WL) test \cite{1968A} is one of the most widely used algorithms for testing graph isomorphism with linear computation complexity \cite{kriege2020survey}. Specifically, the 1-WL algorithm first augments the center node label through aggregating the labels of neighbour nodes as a multiset, and then compresses the augmented labels to new labels \cite{shervashidze2011weisfeiler}. The procedures repeat until the node sets of two graphs differ or the number of repetitions reaches a predetermined value. Although 1-WL works well on testing isomorphic on many graphs \cite{babai1979canonical}, the distinguishing power of the 1-WL algorithm is limited, i.e., 1-WL cannot distinguish any two different $k$-regular graphs of the same order \cite{grohe2017descriptive}, and many other non-regular non-isomorphic graphs \cite{sato2020survey}.

\section{Twin Weisfeiler-Lehman Test}
We propose a novel Twin Weisfeiler-Lehman paradigm for graph isomorphism test, which serves as a theoretical basis for achieving high expressive GNNs beyond 1-WL. Then, theoretical analyses of distinguishing power are presented.

\subsection{Twin Weisfeiler-Lehman Paradigm}
The core of the proposed Twin-WL paradigm lies in the twin message passing process of node labels and node identities. That is, Twin-WL simultaneously passes node label and node identity rather than only pass node label as 1-WL. The identity passing mechanism encodes complete structural information of the rooted subgraph, hence offering additional power beyond the 1-WL at distinguishing graph structures. The twin message passing process proceeds in iterations indexed by $h$ and we describe each iteration detailedly in the following. 

Each iteration consists of \textbf{four steps}, which are multisets determination, multisets sorting, label compression, and relabeling. Specifically, given two graphs $G$ and $G'$, for node $v$, the label is denoted as $l_h(v)$ and the identity is denoted as $id(v)$. In \textbf{step 1}, Twin-WL aggregates the labels and identity sets of neighbor nodes as multisets respectively. Node labels of neighbor nodes are aggregated as a multiset $M_h(v)$. For $h=0$, $M_0(v)=l_{0}(v)$, and for $h>0$, $M_h(v)=\left\{\{l_{h-1}(u)\bm{|}u\in \bm{N}(v) \}\right\}$, where $\bm{N}(v)$ denotes the neighbor nodes of $v$ and $\{\{\}\}$ denotes the multiset. Identity multisets of neighbor nodes are aggregated and combined with the identity of center node which forms a new multiset $t_h(v)$. For $h=0$, $t_{0}(v)= \{\{id(v)\}\}$, and for $h>0$, $t_{h}(v)= \left\{\{id(v), t_{h-1}(u)|u\in \bm{N}(v) \right\}\}$. 
In \textbf{step 2}, each label multiset $M_h(v)$ is sorted and converted to a string $s_h(v)$ with the prefix $l_{h-1}(v)$, which prepares for the label compression. In \textbf{step 3}, each string is compressed to a new label with a hash function $\bm{g}:\sum* \rightarrow \sum$ and $\bm{g}$ should be an injective function. With the sorting step (step 2) above, a simple implementation of $\bm{g}$ is a counter mapping. There is a counter variable $x$ with the initial value of $0$ that records the number of the distinct strings. The variable is incremented by $1$ when a new string is encountered. The mapping alphabet is shared across graphs, which guarantees a common feature space. In \textbf{step 4}, we relabel each node in graph $G$ and $G'$ as $l_{h}(v):=\bm{g}(s_h(v))$. 
The algorithm terminates in $h$-th iteration if:
\begin{equation}
\label{eq_terminateTWL}
 \{(l_h(v),\bm{f}(t_h(v)))\bm{|}v\in V\}\neq \{(l_h(v'),\bm{f}(t_h(v')))\bm{|}v'\in V' \}.     
\end{equation}
To render the identity transferable and comparable across different graphs, a function $\bm{f}$ is required to convert the identity multiset (i.e., identity subtree) to the structural information that can be compared across different graphs. The termination condition means that, if the set that composed of the tuples $(l_h(v),\bm{f}(t_h(v)))$ differs, graph $G$ and $G'$ are determined non-isomorphic. 

The Twin-WL paradigm can be implemented to obtain the graph representations as well, which is analogous to the definition of the feature mapping $\phi$ in the graph kernel methods \cite{borgwardt2005shortest,shervashidze2009efficient,shervashidze2011weisfeiler}. In graph kernel methods, the value of the $i$-th position of the $\phi$ is the number of nodes with the according label. In Twin-WL paradigm, we extend the dimension of the feature representation to form a matrix, and the value in position $(i,j)$ is: 
\begin{equation}
\footnotesize
\begin{split}
&\phi_h^{(i,j)}(G)=\\ &\left\vert \{(l_h(v),\bm{g'}(d_h(v)))\bm{|}l_h(v)=i,\bm{g'}\left(d_h(v)\right)=j,v\in V\} \right\vert,
\end{split}
\end{equation}
where $d_h(v)=\bm{f}(t_h(v))$ is the result of the readout function, $\bm{g'}$ is a counter mapping that counts the number of the different results of the readout function. The final representation of graph is the concatenation of the representations of each layer.

\begin{figure*}[t]
\vskip 0.1in
\begin{center}
\centerline{\includegraphics[width=2\columnwidth]{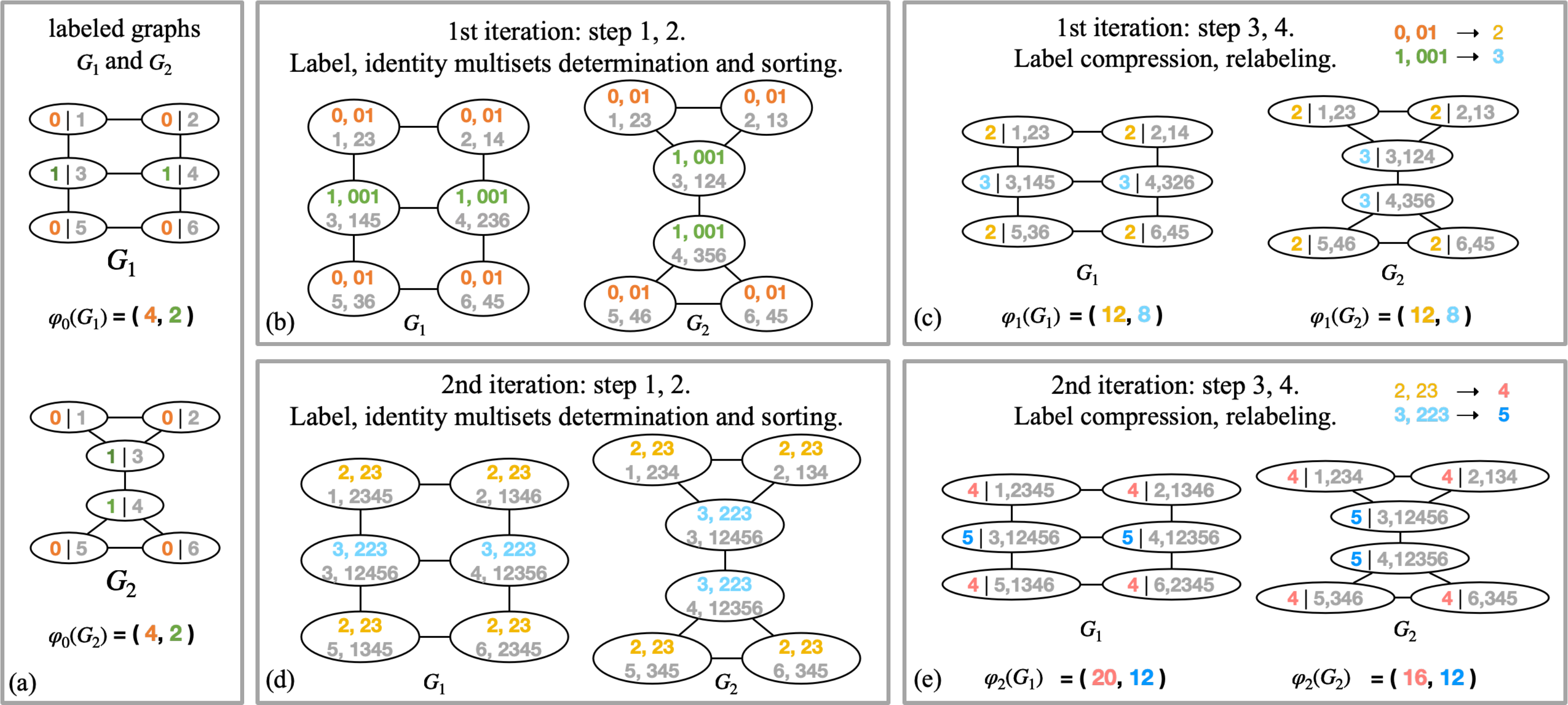}}
\caption{Illustration of the graph representation computation in STwin-GNN. Colored numbers indicate node labels, gray numbers indicate node identities. Node labels and identities are aggregated as multiset and set respectively. The graph representation vector is denoted as $\varphi$, value of the $i$-th position of $\varphi$ is the summation of identity sets sizes with the corresponding label in the tuple. }

\label{fig:algorithm}
\end{center}
\vskip -0.1in
\end{figure*}

\subsection{Distinguishing Power}
Distinguishing power of any implementation of the Twin-WL paradigm is higher than 1-WL for graph isomorphism test. We formalize the statement in Theorem 1 and we further formalize the distinguishing power of any implementation of our Twin-WL paradigm in Theorem 2.    

\paragraph{Theorem 1.}\label{th:th1} \textit{Given two graphs, if they can be distinguished by 1-WL, then they must be distinguished by any implemented algorithm of the proposed Twin-WL paradigm.}

The termination condition of the 1-WL can be denoted as $\{ l_h(v)\bm{|}v\in V\}\neq\{l_h(v')\bm{|}v'\in V'\}$. The termination condition of the implemented method of Twin-WL paradigm is $\{(l_h(v),\bm{f}(t_h(v)))\bm{|}v\in V\}\neq \{(l_h(v'),\bm{f}(t_h(v')))\bm{|}v'\in V' \}$, which is stricter than that of 1-WL by adding a new structural constraint. Therefore, once the graphs are determined unequal by the 1-WL algorithm, they must be determined unequal by the implementation of the Twin-WL paradigm as well.

\paragraph{Theorem 2.} \textit{Given two non-isomorphic graphs $\mathcal{G}$ and $\mathcal{G'}$ that cannot be distinguished by 1-WL, the $h$-hop rooted subgraph centered in node $v\in V(\mathcal{G})$ is denoted as $\mathcal{G}_v^h$, and that of $v'\in V(\mathcal{G'})$ is $\mathcal{G'}_v^h$. Once the encoding results of the defined readout function $\bm{f}$ on $\mathcal{G}_v^h$ and $\mathcal{G'}_v^h$ differ, the implemented method of the Twin-WL paradigm can decide that the two graphs are non-isomorphic.}

The identity multiset $t_h(v)$ in the proposed Twin-WL paradigm encodes the complete structure of the rooted subgraph $\mathcal{G}_v^h$. Once the encoding results of the rooted subgraphs differ, the termination condition of the Twin-WL is satisfied, i.e., $\{(l_h(v),\bm{f}(t_h(v)))\bm{|}v\in V\}\neq \{(l_h(v'),\bm{f}(t_h(v))\bm{|}v'\in V'\}$. Then $\mathcal{G}$ and $\mathcal{G'}$ can be determined non-isomorphic.

\section{Twin Graph Neural Networks}
Based on the proposed Twin-WL paradigm, we provide two graph neural networks that serve different scenarios of graph classification by designing different readout functions. 
\subsection{STwin-Graph Neural Networks }
When the labeled data is insufficient, a method that can be applied and transferred at a low cost is required. Therefore, we choose an inductive metric function that outputs the \underline{s}ize of rooted subgraph as the readout function. The according model is named STwin-GNN. We detailedly describe the model and discuss its expressiveness and complexity.
\subsubsection{Model}
The iteration steps are the same as the Twin-WL paradigm, where the readout function $\bm{f}$ is a metric function of counting the number of node identities within rooted subgraph. In practice, we adopt an equivalent way to realize the above STwin-GNN for a lower computation cost. Specifically, in message passing step (Step 1), we directly pass the identities of neighbor nodes to form an identity set, rather than record 
\begin{algorithm}[H]
\caption{ STwin-GNN for Graph Classification}
\label{algorithm 1}
\KwIn{Graph $G$}

\begin{algorithmic}
\FOR{$h=1$ {\bfseries to} $H$}
\STATE 1. Label multisets and identity sets determination 
\begin{itemize}
    \item[$\bullet$]Aggeregate labels of neighbor nodes centered in each node $v$ in graph $G$ as multiset $M_h(v)$. For $h=0$, $M_0(v)=l_{0}(v)$, for $h>0$, $M_h(v)=\left\{\{l_{h-1}(u)\bm{|}u\in \bm{N}(v) \}\right\}$.
    \item[$\bullet$]Aggregate identity sets of neighbor nodes centered in each node $v$ in graph $G$. Identity of node $v$ and elements in identity sets of neighbor nodes compose the new identity set. For $h=0$, $t_{0}(v)=  \{id(v)\}$, for $h>0$, $t_{h}(v)= \left\{id(v), id(w)\bm{|}w\in t_{h-1}(u),u\in \bm{N}(v) \right\}$.
    
\end{itemize}

\STATE 2. Sorting labels in each label multiset
\begin{itemize}
    \item[$\bullet$]Sort label elements in the label multiset in ascending order and concatenate them into a string $s_h(v)$.
    \item[$\bullet$]Add $l_{h-1}(v)$ as a prefix to $s_h(v)$.
\end{itemize}
\STATE 3. Label compression
\begin{itemize}
    \item[$\bullet$]Map each string $s_h(v)$ to a compressed label using a hash function $g:\sum* \rightarrow \sum$ such that $g(s_h(v)):=g(s_h(w))$ if and only if $s_h(v)=s_h(w)$.
\end{itemize}
\STATE 4. Relabeling
\begin{itemize}
    \item[$\bullet$] Set $l_{h}(v):=g(s_h(v))$ for all nodes in $G$.
\end{itemize}
\STATE $\phi_h(G)=\sum |\{(l_h(v),t_h(v)),l_{h}(v)=h\} |$
\ENDFOR

\end{algorithmic}
\KwOut{ $\phi(G)=\left[\phi_0(G),...,\phi_h(G)\right]$}

\end{algorithm}
 the entire identity subtree structure, e.g., $t_{h}(v)= \left\{id(v), id(w)\bm{|}w\in t_{h-1}(u),u\in \bm{N}(v) \right\}$. Such identity set is sufficient as well as efficient for counting the total number of node identities within rooted subgraph. In other words, a counting function $\bm{f'}$ that counts the size of identity set $t_h(v)$ is adopted, i.e., $\bm{f'}=|t_h(v)|$. We show the steps of the STwin-GNN in Algorithm \ref{algorithm 1}.

In order to avoid dimension disaster and the sparsity of representation in the specific scenario, we adopt an alternative version of formula (2). The value of the $i$-th position of graph representation $\phi$ is:

\begin{equation}
    \phi_h^{(i)}(G)=\sum_{l_h(v)=i,v\in V}\left|t_h(v)\right|,    
\end{equation}
which means the summation of the identity set size of nodes with the same label. We illustrate the two iterations of the STwin-GNN in Figure \ref{fig:algorithm} (a)-(e), where colored numbers indicate node labels, gray numbers indicate node identities. The iteration of the only colored number is the process of the 1-WL, which cannot discriminate the graph $G_1$ and $G_2$. Take the result of the 2nd iteration as an example, the representations of $G$ and $G'$ obtained by 1-WL are the same, i.e., $\phi_2(G_1)=\phi_2(G_2)=(4,2)$. As for our STwin-GNN, $\phi_2(G_1)=(\sum_{l_h(v)=4}(|\{1,2,3,4,5\}|+|\{2,1,3,4,6\}|+|\{5,1,3,4,6\}|+|\{6,2,3,4,5\}|),...)=(20,12)$, while  $\phi_2(G_2)=(\sum_{l_h(v)=4}(|\{1,2,3,4\}|+|\{2,1,3,4\}|+|\{5,3,4,6\}|+|\{6,3,4,5\}|),...)=(16,12)$. STwin-GNN can discriminate graph $G_1$ and $G_2$. The graph representations obtained by STwin-GNN is more distinguishable than that of 1-WL.

Finally, the outputs of each layer are concatenated as the final graph representation, $\phi(G)=[\phi_1^{(1)},..., \phi_1^{(i)},...,\phi_h^{(1)},..., \phi_h^{(i)}]$, e.g., the representation of $G_1$ in Figure \ref{fig:algorithm} is $\phi(G_1)=(4, 2, 12, 8, 20, 12)$. Multilayer perceptrons (MLP) are utilized as the classifier, and the obtained graph representations are sent to the classifier to get the classification results. The cross entropy loss is adopted to optimize the model.

\subsubsection{Disscussion}
\paragraph{Expressive Power} As stated in Theorem 1, the termination condition of the STwin-GNN algorithm is stricter than that of 1-WL. Therefore, once the graphs are determined unequal by the 1-WL algorithm, they must be determined unequal by STwin-GNN as well. In addition, once the nodes numbers of the rooted subgraphs $\mathcal{G}_v^h$ and $\mathcal{G'}_v^h$ differ, the STwin-GNN can decide that the two graphs are non-isomorphic. The expressive power of the STwin-GNN is higher than that of GNNs based on the 1-WL.

\begin{figure*}[ht]
\vskip 0.1in
\begin{center}
\centerline{\includegraphics[width=2\columnwidth]{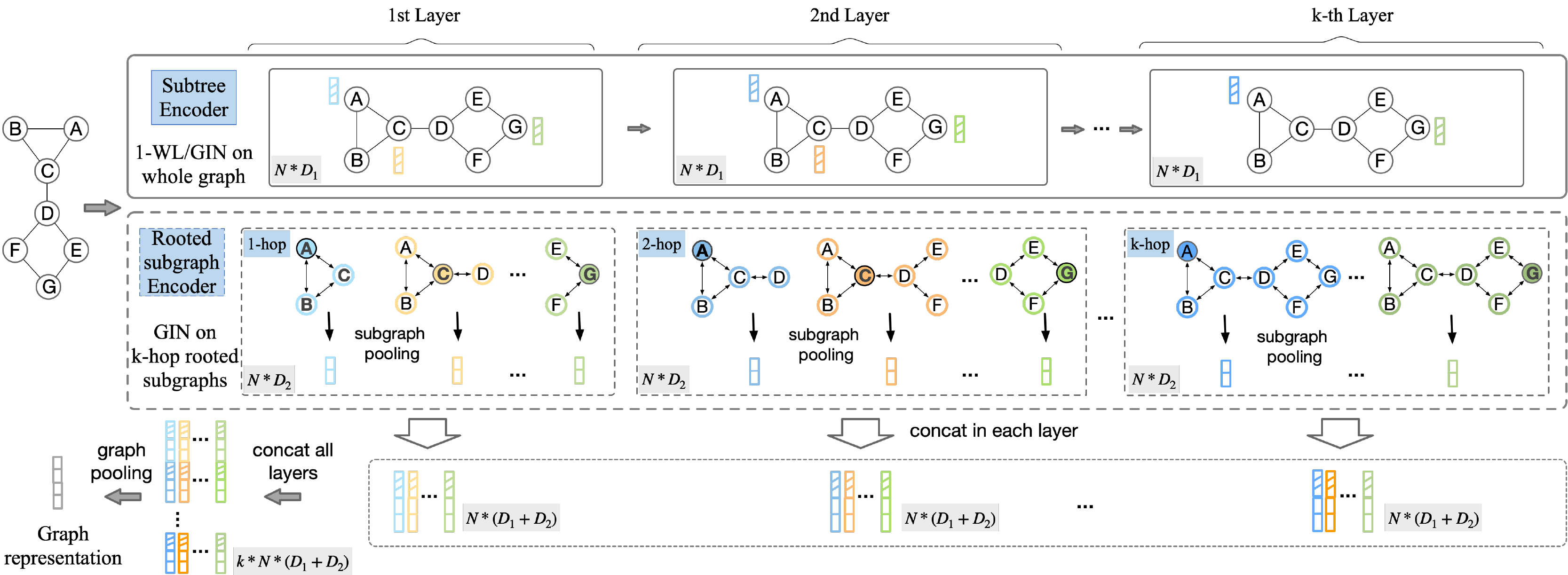}}
\caption{NTwin-GNN consits of a subtree encoder and a rooted subgraph encoder. The input of the subtree encoder is the whole graph. For the rooted subgraph encoder, the input of $k$-th layer are $k$-hop rooted subgraphs centered in each node. Then sum pooling is performed to obtain the representation of the center node. The outputs of two encoders are concatenated as the nodes representations. The final graph representation is obtained by sum pooling. Bars with the same color denote representations of the same node.}
\label{fig:model}
\end{center}
\vskip -0.1in
\end{figure*}

\paragraph{Complexity}
Given the graph $G$ with node number $N$, average node degree $D$ and edge number $M$, where $M=ND$. The time complexity of the STwin-GNN with $h$ iterations is $O(hM)$. 
In each iteration, the STwin-GNN mainly contains four step. In step 1, determining the label multisets and identity sets for all nodes takes $O(ND)$ operations which can be accomplished simultaneously. The runtime of the identity set can be achieved by using a hash table. In step 2, The complexity of sorting the multiset is $O(ND)$ which can be achieved by using counting sort. In step 3, the label compression requires passing over all strings and it takes $O(ND)$. Therefore, these steps of graph isomorphism determination take a total runtime of $O(hND)$ for $h$ iterations. The total time complexity is $O(hND)$, which equals the time complexity of the 1-WL algorithm $O(hM)$ \cite{shervashidze2011weisfeiler}.

\subsection{NTwin Graph Neural Network}  
When the labeled data is sufficient, we choose more powerful \underline{n}eural model as readout function that outputs the abundant structural information of the rooted subgraph. The main advantage of the neural implementation is that any neural model with high expressiveness can be chosen as readout function. Besides, any implementation with neural model can be more expressive than 1-WL. We detailedly describe one implementation named NTwin-GNN and discuss its expressiveness and complexity.

\subsubsection{Model}
Following the Twin-WL, NTwin-GNN composes of two encoders with respective purposes as well. One is the subtree encoder to encode the node labels. The other is the rooted subgraph encoder to read out the structural information around a node. In each layer, the two components work separately and the results of them are concatenated as the output of the layer. We detail each component in the following and illustrate the NTwin-GNN in Figure \ref{fig:model}.

\paragraph{Subtree Encoder} The subtree encoder is designed to preserve the label subtree of each node. Since the compressed node label in 1-WL algorithm represents the according subtree pattern \cite{shervashidze2011weisfeiler}, we adopt the 1-WL algorithm \cite{1968A} running on the whole graph as the subtree encoder. Given a set of graphs, we assume one of the graphs is denoted as $G=(V,E)$, where $v\in \mathbb{R}^{N\times D}$, $N$ is the number of nodes, $D$ is the dimension of feature vector. For each node $v\in V(G)$, the subtree encoder augments the node labels by the sorted set of neighbor nodes and compresses the augmented labels to new labels. The detailed description of the 1-WL algorithm can be found in \cite{shervashidze2011weisfeiler}. Each layer in the encoder corresponds to one iteration in 1-WL, and we take the compressed node label in $k$-th layer as the node representation. Initial node label of $v$ is denoted as $l_v^{(0)}$, and the compressed node label in $k$-th layer is $l_v^{(k)}$. Then the output of the $k$-th layer subtree encoder is : 
\begin{equation}
    \bm{g}_v^{(k)} = \bm{s}_{v}^{(k)}
\end{equation}
where $\bm{g}^{(k)}_v\in \mathbb{R}^{D_1}$ is the representation of node $v$ in the $k$-th layer and $D_1$ is the total number of node labels in the layer. $\bm{s}_{v}^{(k)}$ is the onehot vector of node label $l_v^{(k)}$. 
In the $k$-th layer, each node label represents a subtree pattern with height $k$. Note that in the cases of continuous node features, 1-WL algorithm in the subtree encoder can be substituted with Graph Isomorphism Network (GIN) \cite{xu2018powerful}.

\paragraph{Rooted subgraph Encoder} 
The rooted subgraph encoder is designed to read out the structural information of the rooted subgraph. Specifically, we adopt GIN \cite{xu2018powerful} with $\epsilon=0$ to obtain the representation of each node in the rooted subgraph, then leverage the sum pooling function over the rooted subgraph to obtain the structural representation of the node. Given a graph $G=(V,E)$, number of nodes is $|V(G)|=N$. In the $k$-th layer, the inputs of rooted subgraph encoder are $k$-hop rooted subgraphs $\mathcal{G}_v^k$ centered in each node $v\in V(G)$. GIN is performed: 
\begin{equation}
    \bm{h}_p^{(k)}=\mathrm{MLP}^{(k)}\left(\boldsymbol{h}_p^{(k-1)}+\sum_{q\in \bm{N}(p)} \boldsymbol{h}_{q}^{(k-1)}\right),
\end{equation}
where $p$ denotes node in rooted subgraph $\mathcal{G}_v^k$. For each rooted subgraph, GIN aggregates features of the neighbor nodes, which are then combined with the center node and the combined features are mapped to new features. Then, sum pooling is performed to obtain the whole rooted subgraph representation:
\begin{equation}
    \bm{h}_{v}^{(k)}=\mathrm{SUM}\left(\{\bm{h}_p^{(k)} |p\in V(\mathcal{G}_v^k)\}\right),
\end{equation}
where $v$ denotes node in graph $G$ and $\bm{h}_{v}^{(k)}\in \mathbb{R}^{D_2}$. The representation of subgraph rooted in node $v$ is regard as the structural feature of node $v$ in the graph $G$.

\paragraph{Graph Representation} Then, in the $k$-th layer, we concatenate the outputs of the subtree encoder and the rooted subgraph encoder for all nodes as the node representations of the $k$-th layer. Representataion of one node in $k$-th layer is $\bm{h}^{(k)}(v) = [\bm{g}_v^{(k)},\bm{h}_v^{(k)}] \in \mathbb{R}^{(D_1+D_2)}$. The graph representation in the $k$-th layer is obtained by sum pooling, i.e., $\bm{H}^{(k)}(G)=\mathrm{SUM}\left(\bm{h}^{(k)}(v)|v\in V(G) \right)$ and $\bm{H}^{(k)}(G)\in  \mathbb{R}^{(D_1+D_2)}$. Then, outputs of all $k$ layers are concatenated as the final graph representation $H(G)=[\bm{H}^{(1)}(G),\bm{H}^{(2)}(G),...,\bm{H}^{(k)}(G)]\in  \mathbb{R}^{(D_1+D_2)*k}$. 

\subsubsection{Discussion}

\paragraph{Expressive Power}
Compared with the graph neural networks that directly aggregate node labels over the whole graph, our NTwin-GNN not only encode the node labels, but also encode the structure of the rooted subgraph centered in the node, which is significant for the graph classification task. The expressive power of NTwin-GNN is higher than any standard GNNs based on the 1-WL algorithm.
\vspace{0.1in}
\paragraph{Complexity} We analyse the time complexity of NTwin-GNN. Given a graph with $N$ nodes, the average degree of nodes is $D$ and the max nodes number of rooted subgraphs is $n$. The subtree encoder works on whole graphs takes $O(ND)$ operations. The rooted subgraph encoder takes $O(NnD)$, where $n$ can be small with few hops of rooted subgraph. The whole time complexity is $O(ND+NnD)$.

\section{Experiments}

\begin{table*}[t]
\caption{10-Fold Cross Validation average test accuracy and standard deviation on TU datasets.}
\label{TU}
\centering
\vskip 0.15in
\resizebox{0.9\textwidth}{!}{
\begin{sc}
\begin{tabular}{l|cccccc}
\toprule

Methods &   MUTAG &PTC\_MR & Mutagenicity & NCI1 & NCI109 \\ 
\midrule
SP kernel  & $87.28\pm0.55$ & $58.24\pm2.44$ & $71.63\pm2.19$ & $73.47\pm0.21$ & $73.07\pm0.11$   \\ 
WL kernel  & $82.05\pm0.36$ & $57.97\pm0.49$ & - & $82.19\pm0.18$ & $82.46\pm0.24$   \\ 
DGK & $87.44\pm2.72$ & $60.08\pm2.55$ & - & $73.55\pm0.51$  & $73.26\pm0.26$    \\   
\midrule
GCN  & $78.69\pm6.56$ &$66.73\pm4.65         $& $80.84\pm1.35$ & $78.39\pm1.79$ &  $77.57\pm1.79 $\\
GIN  & $81.51\pm 8.47$  & $54.09\pm 6.20$ & $77.70\pm 2.50$  & $80.0\pm1.40$ & $70.20\pm3.21$ \\
Diffpool & $80.00\pm6.98$ & $57.14\pm7.11$	 &$80.55\pm1.98$   & $78.88\pm3.05$ &  $76.76\pm2.38$   \\ 
SortPool &  $85.83\pm1.66$ & $58.59\pm2.47$ 	 & $80.41\pm1.02$  & $74.44\pm0.47$  & -    \\ 
\midrule
1-2-3-GNN    & $86.10\pm0.0$ & $60.9\pm0.0$& -&  $76.2\pm0.0$ &-         \\
3-hop GNN      & $87.56\pm0.72$ &-& - & $80.61\pm0.34$ & - \\
Nested GIN  & $87.9\pm8.2$ &$54.1\pm7.7$&$82.4\pm2.0$ & $78.60\pm2.30$ &$77.2\pm2.9$ \\
\midrule
\textbf{STwin-GNN}  & $\mathbf{90.00\pm3.89}$&  $\mathbf{70.33\pm5.32}$    & $ \mathbf{84.32\pm1.48} $ & $\mathbf{84.45\pm0.66}$ & $\mathbf{85.37\pm0.81}$\\
\textbf{NTwin-GNN}    & $\mathbf{88.89\pm3.74}$&  $\mathbf{72.22\pm5.36}$ & $\mathbf{84.63\pm3.74} $ & $\mathbf{85.55\pm0.97}$ & $\mathbf{85.64\pm0.83}$ \\
\bottomrule
\end{tabular}
\end{sc}
}
\vskip -0.1in
\end{table*}

In this section, we evaluate the effectiveness and the expressive power of our STwin-GNN and NTwin-GNN. We first evaluate the performance of Twin-GNNs on the graph classification task to verify the effectiveness. Then, we conduct experiment to verify that the expressive power of Twin-GNNs is strictly higher than that of 1-WL algorithm.

\subsection{Datasets} 
Performance of STwin-GNN and NTwin-GNN on graph classification task are evaluated on benchmark datasets including: MUTAG \cite{debnath1991structure}, PTC\_MR \cite{toivonen2003statistical}, Mutagenicity \cite{kazius2005derivation}, NCI1 \cite{wale2008comparison} and NCI109 \cite{wale2008comparison}. Graphs in these datasets represent chemical molecules, nodes represent atoms and the edges represent chemical bond. We provide detailed descriptions and statistics of the above datasets in the Appendix. 
The expressive power of STwin-GNN and NTwin-GNN are evaluated on the EXP dataset \cite{ACGL-IJCAI21}. EXP dataset contains 600 pairs of graphs that are non-isomorphic and are 1-WL indistinguishable.

\subsection{Baselines}
In the experiment of the graph classification task, we adopt three graph kernel methods, some GNNs methods based on the 1-WL, and some methods with higher expressive power than 1-WL as baselines. Graph kernel methods which include shortest path kernel \cite{borgwardt2005shortest}, WL subtree kernel \cite{shervashidze2011weisfeiler} and deep graph kernel \cite{yanardag2015deep}. GNNs methods based on the 1-WL include GCN \cite{kipf2017semi}, GIN \cite{xu2018powerful}, Diffpool\cite{ying2018hierarchical}, and Sortpool \cite{lee2019self}. For GCN, graph representations are obtained by the learned nodes representations and sum pooling. Higher expressive methods include 1-2-3 GNN \cite{morris2019weisfeiler}, 3-hop GNN \cite{nikolentzos2020k} and the Nested GNN \cite{zhang2021nested}. Results of baselines are obtained either from raw paper or source code with published experimental settings ("-" indicates that results are not available). For GCN and GIN , we search the model layer in $\{2,3,4,5\}$, and hidden dimensions in $\{32, 64, 128\}$. For Nested GNN, we choose the best-performing Nested GIN as baseline according to the results in the original paper. On the datasets Mutagenicity, NCI and NCI109, we search the subgraph height in $\{2,3,4,5\}$ with 4 model layers. 
In the experiments of expressive power evaluation, we adopt GCN, GIN, PPNG \cite{maron2019provably}, and GCN-RNI \cite{ACGL-IJCAI21} as baselines. GCN and GIN represent neural versions of the 1-WL, and the expressive power of the two methods is at most as large as 1-WL. PPGN is a high order GNNs with higher expressive power. GCN-RNI is a GCN model with random node initialization.

\subsection{Experimental Setup}
We perform 10-fold cross validation where 9 folds for training, 1 fold for testing. $10\%$ split of the training set is used for model selection \cite{errica2019fair}. We report the average and standard deviation (in percentage) of test accuracy across the 10 folds. We implement experiments with PyTorch and employ Adam optimizer with the learning rate as 0.001 to optimize the model. For our STwin-GNN and NTwin-GNNs, we search the hop number of rooted subgraphs and model layer in $\{2,3,4,5\}$ respectively. We train the models with batch size 32. We take sum pooling as subgraph pooling and the graph pooling uniformly. In the training process, we set the maximum number of iterations 100 and adopt early stopping with patience 15. For a fair comparison, the MLP used as classifier are 2 layers with hidden dimension 64. 

\subsection{Performance on Graph Classification Task}
Results of the graph classification are shown in Table \ref{TU}. Compared with graph kernel methods, our STwin-GNN and NTwin-GNN gain strong improvements in the all TU datasets. Especially, both Twin-GNNs achieve better performance than WL subtree kernel which proves the higher discriminative power experimentally. It verifies that the augmented structural information of the rooted subgraph on the basis of the subtree is effective on the graph classification task. Compared to the standard GNNs based on 1-WL, i.e., GCN, GIN, Diffpool and Sortpool, both Twin-GNNs consistently outperforms these methods significantly. The improvements demonstrate that the structural features obtained by our methods are more effective than those obtained by standard GNNs and pooling strategies. For 1-2-3-GNN, 3-hop GNN and Nested GIN, it has been proved that the representation power is higher than standard message passing GNNs in their original papers. Our Twin-GNNs still outperforms the three methods in all datasets except on which the results are unavailable. Especially, our STwin-GNN gain such progress with low computational cost. In addition, compared with GIN, the improvements verify the effectiveness of the rooted subgraph encoder in NTwin-GNN.

\subsection{Expressive Power Evaluation}
Results on the EXP dataset are showed in Table \ref{exp}. EXP is a synthetic dataset which is constructed for the expressive power evaluation \cite{ACGL-IJCAI21}. Each pair graph in EXP is non-isomorphic and 1-WL indistinguishable, and it can be classified correctly by high order GNNs. The results demonstrate that GNNs based on 1-WL cannot distinguish any pair of the graphs, i.e., GCN and GIN. Despite the expressive of 2-WL, the accuracy of PPNG is similar to that of GCN. Some results of baselines are from \cite{ACGL-IJCAI21}. While for our methods, STwin-GNN and NTwin-GNN consistently achieve very high accuracy, which can distinguish nearly all graph pairs. The results verify the high expressive power of our implementations of the Twin-WL which is stated theoretically in section 3.2. 

\subsection{Case Study}
\begin{figure}[ht]
\vskip 0.1in
\begin{center}
\centerline{\includegraphics[width=1\columnwidth]{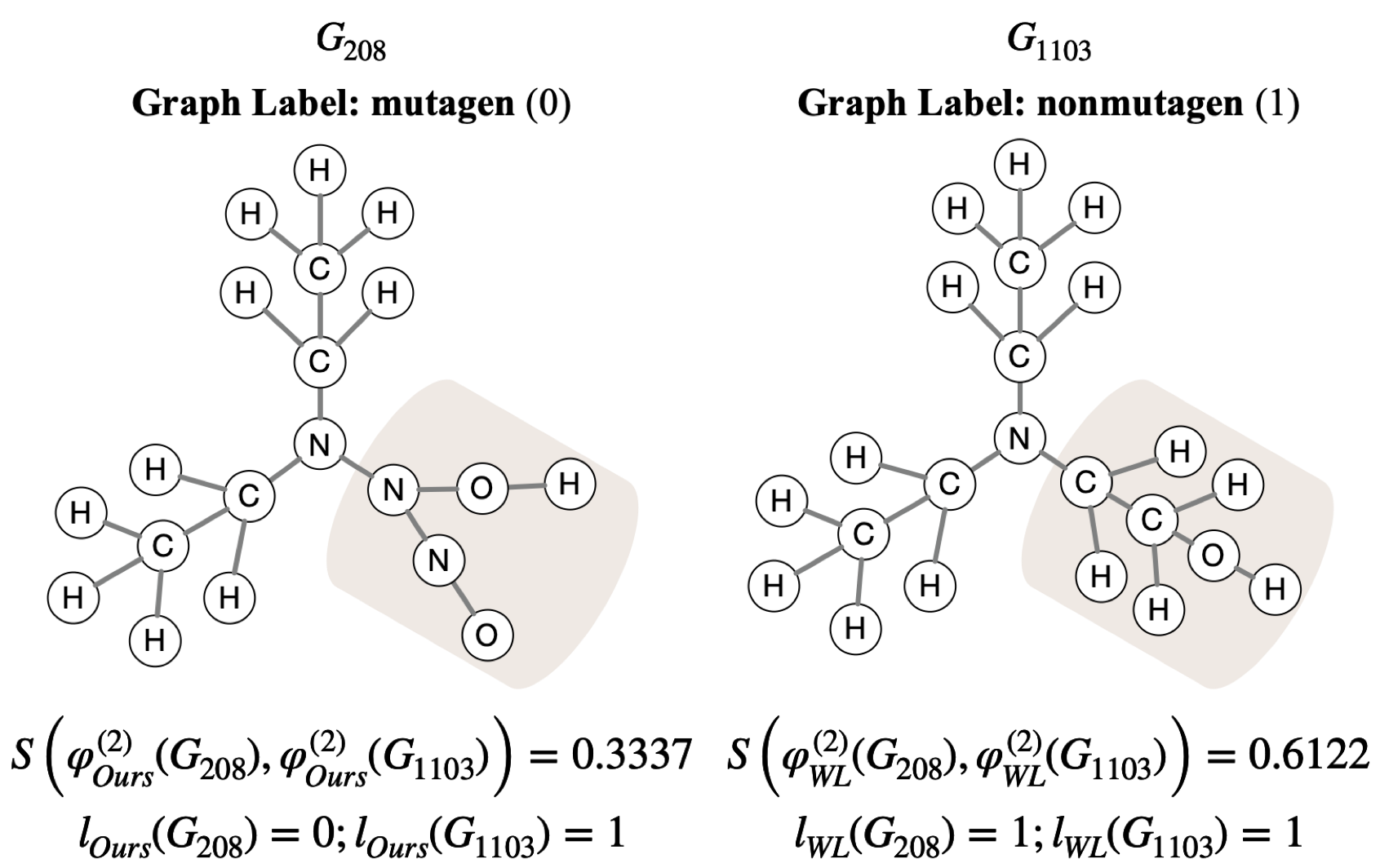}}
\caption{Real cases from Mutagenicity. Letters in node denote the node label. Shaded areas are the difference of the $G_{208}$ and $G_{1103}$.} 
\label{fig:case}
\end{center}
\vskip -0.1in
\end{figure}

In order to intuitively show that the distinguishing power of STwin-GNN is higher than that of 1-WL, we provide two real cases in dataset Mutagenicity. $G_{208}$ is the graph of index 208 with graph label mutagen. $G_{1103}$ is the graph of index 1103 with graph label nonmutagen. We first compute the graph representations by STwin-GNN and 1-WL respectively, then compute the cosine similarity of the representations. 
We illustrate the two graphs in Figure~\ref{fig:case}. $\varphi_{Ours}^{(2)}$ and $l_{Ours}$ denote graph representations and classified results obtained by our STwin-GNN with two iterations respectively. $S(a,b)$ denotes the cosine similarity of $a$ and $b$. 
STwin-GNN classifies the two graphs correctly as different labels, while the 1-WL classifies the two graph as the same label. In addition, the similarity of the representations obtained by STwin-GNN is lower than that of 1-WL algorithm. The facts experimentally demonstrate that the discriminate power of STwin-GNN is higher than that of 1-WL.

\begin{table}[t]
\caption{Expressive Power Evaluation on EXP dataset.}
\label{exp}
\vskip 0.1in
\begin{center}
\begin{small}
\begin{sc}
\begin{tabular}{lc}
\toprule
Model & Test Accuracy (\%) \\
\midrule
GCN \cite{kipf2017semi} & $50.0\pm 0.00$ \\
GIN \cite{xu2018powerful}     & $50\pm 0.00$  \\
PPNG \cite{maron2019provably}   & $50.0\pm 0.00$ \\
GCN-RNI \cite{ACGL-IJCAI21}   &  $98.0\pm 1.85$\\
\midrule
\textbf{STwin-GNN}      & $\mathbf{99.50\pm 0.70}$ \\
\textbf{NTwin-GNN}   & $\mathbf{99.33\pm0.93}$ \\
\bottomrule
\end{tabular}
\end{sc}
\end{small}
\end{center}
\vskip -0.1in
\end{table}

\subsection{Runtime Comparison}
Our STwin-GNN has higher discriminative power than 1-WL with the same time complexity theoretically. In order to compare the time cost of the two methods practically, we record their running time in obtaining representations of all graphs in three datasets respectively. We show the mean runtime (second) and standard deviation comparison in Tabel~\ref{runtime}. We run each method ten times on each dataset and conduct t-test as a significance test. The p-value is $0.1722 > 0.05$, which demonstrate that no significant difference in runtime of STwin-GNN and 1-WL.

\section{Related Works}

\textbf{Methods for Graph Isomorphism Test}
The Graph Isomorphism (GI) problem is the algorithmic problem to decide whether two graphs are structurally identical \cite{grohe2020graph}. It has been proved that the GI problem can be solved in quasipolynomial (exp(($\log n^{O(1)}$))) time theoretically \cite{babai2016graph}. Weisferiler and Lehman (WL) algorithm is one of the simplest approaches to the GI problem which can work well on many graphs \cite{grohe2017descriptive}. However, 1-WL is limited in distinguishing $k$-regular graphs and some non-isomorphic graph pairs. Subsequently, 1-WL is generalized to high dimensions as $k$-WL algorithm. However, $k$-WL takes $k$-nodes tuples as atomic facts, and the enumeration of the tuples makes it a computationally complex method. The runtime complexity is $O(n^{k+1}\log n)$. Different from $k$-WL, our Twin-WL enhance the distinguishing power on the basis of subtree pattern obtained by 1-WL, and it can be achieved synchronously during the 1-WL iterations, no additional runtime is required.

\begin{table}[t]
\caption{Runtime Comparison (second).}
\label{runtime}
\centering
\vskip 0.15in
\resizebox{0.5\textwidth}{!}{
\begin{sc}
\begin{tabular}{l|cccccc}
\toprule
Model & Mutagenicity & NCI1& NCI109  \\
\midrule
STwin-GNN & $4.99\pm0.22$ &$4.81\pm0.20$ &$4.96\pm0.20$ \\
WL     & $4.90\pm0.23$&$4.69\pm0.16$ &$4.73\pm0.20$ \\
\bottomrule
\end{tabular}
\end{sc}
}
\vskip -0.1in
\end{table}

\textbf{Expressive GNNs beyond 1-WL algorithm}
GNNs based on the 1-WL algorithm have been proved at most as powerful as 1-WL with the injective aggregation and pooling functions \cite{xu2018powerful,morris2019weisfeiler}. The fact limits the discriminative power of GNNs. Therefore, more and more kinds of GNNs with higher expressive power than 1-WL have been proposed recently. One intuitive idea is to build GNNs based on high-dimension WL algorithm, e.g., PPNG \cite{maron2019provably} based on the 2-WL algorithm, $k$-GNNs \cite{morris2019weisfeiler} based on set $k$-WL algorithm. However, the high dimension WL algorithms require enumeration of the nodes tuple, which limits the scalability and generalization with high computational cost. Then, many methods intend to improve the expressive power of GNNs, e.g., ID-GNNs \cite{you2021identity}, Nested GNN \cite{zhang2021nested}. More related works are discussed in the Appendix.

\section{Conclusion}
The widely adopted message passing graph neural networks (GNNs) are at most as powerful as 1-WL. We propose Twin-WL paradigm as the theoretical basis for high expressive GNNs beyond 1-WL. The Twin-WL conducts label and identity passing simultaneously without extra runtime. The additional identity passing preserves a complete substructure around a node which offers extra expressiveness. To further readout the structure information, we provide two implementations including STwin-GNN and NTwin-GNN for different scenarios. We experimentally demonstrate the effectiveness and the high expressiveness of our Twin-GNNs.

\newpage



\nocite{langley00}

\bibliography{example_paper}
\bibliographystyle{icml2021}


\end{document}